%% file: ms.tex
\documentclass[11pt]{article} 

\usepackage{graphicx}
\usepackage{booktabs}
\usepackage[T1]{fontenc}
\usepackage{lmodern}
\usepackage[flushleft]{threeparttable}
\usepackage[hidelinks]{hyperref} %
\hypersetup{hidelinks}
\usepackage{multirow}
\usepackage{longtable}
\usepackage[margin=3cm]{geometry} %
\usepackage{amsmath}
\usepackage{authblk}
\usepackage{caption}

\title{Google Coral-based edge computing person reidentification using human parsing combined with analytical method}
\author[1,*]{Nikita Gabdullin}
\author[1,*]{Anton Raskovalov}
\affil[1]{Joint Stock "Research and production company "Kryptonite" \authorcr
E-mail: n.gabdullin@kryptonite.ru, a.raskovalov@kryptonite.ru}
\affil[*]{equal contribtution}
\date{}
\begin{document}

    \captionsetup[figure]{labelformat={default},labelsep=period,name={Figure}}
    \captionsetup[table]{labelformat={default},labelsep=period,name={Table}}

    \maketitle

    \begin{abstract}
        Person reidentification (re-ID) is becoming one of the most significant application areas of computer vision due to its 
        importance for science and social security. Due to enormous size and scale of camera systems it is beneficial to develop edge 
        computing re-ID applications where at least part of the analysis could be performed by the cameras. However, conventional 
        re-ID relies heavily on deep learning (DL) computationally demanding models which are not readily applicable for edge computing. 
        In this paper we adapt a recently proposed re-ID method that combines DL human parsing with analytical feature extraction and 
        ranking schemes to be more suitable for edge computing re-ID. First, we compare parsers that use ResNet101, ResNet18, MobileNetV2,
        and OSNet backbones and show that parsing can be performed using compact backbones with sufficient accuracy. Second, we transfer 
        parsers to tensor processing unit (TPU) of Google Coral Dev Board and show that it can act as a portable edge computing re-ID 
        station. We also implement the analytical part of re-ID method on Coral CPU to ensure that it can perform a complete re-ID cycle. 
        For quantitative analysis we compare inference speed, parsing masks, and re-ID accuracy on GPU and Coral TPU depending on parser 
        backbone. We also discuss possible application scenarios of edge computing in re-ID taking into account known limitations mainly 
        related to memory and storage space of portable devices. 
    \end{abstract}

    
    \input{full}

\end{document}

%% file: full.tex
\section{Introduction}\
\label{introduction}

Computer vision (CV) is a cornerstone of conventional computer science
due to its importance for image and video analysis, robotics, autonomous
vehicles, manufacturing, and others. CV plays a crucial role in video
surveillance which in turn is becoming indispensable for security in
modern cities. Along with well-known topics such as object detection and
face recognition, person re-identification (re-ID) plays an important
role in ensuring public safety by focusing on person identification
across camera systems.

Nowadays the most high-performance re-ID methods rely on deep learning
(DL) \cite{reidrev}. However, there is a known problem of poor
generalization where model's performance decreases when applied to
previously unseen data \cite{pyramid}. Furthermore, large model size
originating from tens to hundreds of millions of parameters associated
with deep learning architectures makes some models unusable when
computational resources and storage space are limited. This complicates
their application in edge computing when some computations are performed
by the camera module \cite{edgerev}. Such applications demand as compact
DL models as possible. This inspired, for instance, the development of
MobileNet \cite{mobilenetp} which is yet to be applied to re-ID tasks.

Recently we proposed re-ID method which combines deep learning-based
human parsing with analytical feature extraction \cite{combreid}. It
showed potential to improve generalization by being dataset-agnostic
relying on DL for human parsing only with most of the analysis conducted
using a simple analytical model. However, the original implementation of
our method relied on a parser that used ResNet-101 backbone \cite{schpp}
which diminished the benefits of low space requirements that the
analytical method provided. In this paper we show that this combined
method can be implemented using a significantly more compact backbone
with little reduction in re-ID accuracy metrics. This allows us to
create an edge computing application which can perform a full re-ID
cycle.

Google Coral Dev Board (Coral) is a single board computer
developed by Google \cite{Coralp}. Coral has a single Edge Tensor
Processing Unit (TPU) processor for neural network operation
acceleration \cite{tpu}. Coral has been shown to have the best
performance for inference time and power consumption among similar edge
computing platforms: Asus Tinker Edge R, Raspberry Pi 4, Google Coral
Dev Board, Nvidia Jetson Nano, and microcontroller Arduino Nano 33 BLE
\cite{perf}.

The motivation for this study is two-fold. First, we aim to show that
our combined method proposed in \cite{combreid} can be implemented with
compact parsers without any significant accuracy loss. Second, we
explore the possibility of transferring parsers based on different
backbones to Google Coral. Finally, full re-ID cycle is performed on
Coral and application scenarios of edge computing re-ID model are
discussed.

The rest of the paper is organized as follows: Section~\ref{methodology} outlines
methodology with respect to parser training and its transfer to Coral,
Section~\ref{experiments} summarizes experimental results, Section~\ref{discussions} provide a
discussion and Section~\ref{conclusions} concludes the paper.

\section{Methodology}
\label{methodology}

\subsection{Human parsing for re-ID}
\label{human-parsing-for-re-id}

Human parsing is subset of image segmentation algorithms which focuses
specifically on humans. It provides parsing maps where every image pixel
is assigned a class specific to human body parts and clothes. Human
parsing is often performed using DL-based models trained on dedicated
datasets such as LIP \cite{lipp} or Pascal \cite{paskp}.

Human parsing was previously used as part of DL-based re-ID models in 
\cite{onesh,hpreid}. However, feature extraction and comparison were
also performed in those studies by neural networks making the complete
model rather large and not particularly suitable for edge computing
applications. To address this issue parser size should be minimized
along with further simplification of the model and performing analytical
feature extraction and similarity ranking will significantly reduce the
total size of the model.

In this study we investigate the possibility of changing the parser
backbone to reduce its size, preferably with minimal loss in accuracy.
We make a number of changes to the parser architecture proposed in
\cite{schpp} which in turn relies on CE2P architecture \cite{ce2pp}. This
parser consists of a backbone and three branches, namely Edge, Decoder,
and Fusion, and only the backbone is changed in our experiments while
the branches remain unchanged. However, the number of parameters in the
branch layers depends on the number of feature channels in backbone's
layers, so it decreases for more compact backbones. The modifications
are discussed in detail in the next subsection.

\begin{table}
  \caption{Accuracy metrics of the combined re-ID method depending on
  parser backbone.}
  \label{tab:1}
  \centering
  \begin{tabular}{|c|c|c|c|c|c|c|c|c|c|} 
    \hline
    \multirow{2}{*}{} & \multicolumn{3}{c|}{Original} & \multicolumn{3}{c|}{Quantized} & \multicolumn{3}{c|}{Coral} \\
    \cline{2-10}
                      & rank-1 & rank-10 & mAP       & rank-1 & rank-10 & mAP       & rank-1 & rank-10 & mAP \\
    \hline
    ResNet101 & \textbf{92.1} & \textbf{97} & 
    \textbf{24} & 90.5 & 96.2 & \textbf{22.1} & 90.2 & 96.5 &
    \textbf{22.1} \\
    \hline
    ResNet18 & 89.7 & 95.4 & 21.8 & 88.3 & 94.7 & 20.3
    & 88 & 94.9 & 17.8 \\
    \hline
    MobileNetV2 & 89.3 & 96.7 & 20.2 & 89.3 &
    95.9 & 20.2 & 88.9 & 96 & 20.1 \\
    \hline
    OSNet & 89.6 & 96.6 & 20.3 & \textbf{91.2} & \textbf{97} &
    21 & \textbf{91.2} & \textbf{96.9} & 21 \\ 
    \hline                
  \end{tabular}
\end{table}

\subsection{Parser modification and its transfer to
Coral}
\label{parser-mod}

To perform inference on Coral a neural network model has to be compiled
with edgetpu compiler. This compilation is possible only for quantized
TensorFlow Lite (tflite) models. To obtain such models the original
SCHP-parser implemented using PyTorch \cite{pytorchp} is converted as
follows. First, it is converted into ONNX format \cite{onnxp}, then to
OpenVINO format \cite{ovinop} and then to Tensorflow using
openvino2tensorflow convertor \cite{ov2tf}. Finally, the obtained model which has
format of ``saved model'' is quantized into tflite model. It should be
noted that we substitute CUDA-dependent InplaceABNSync layers \cite{InAB}
in SCHP parser which are not supported by ONNX converter with BatchNorm
and ReLU. OpenVINO and openvino2tensorflow converters allow
us to change dimensions order in model tensors, which are 
different in PyTorch and Tensorflow formats.

TensorFlow post-training integer quantization requires ``a
representative dataset'' (further referred to as quantization dataset)
for processing. This is necessary to choose quantization coefficients
for better translation of float input data to integer values. We used
random subsets of LIP dataset for quantization, and we discuss the
influence of the quantization dataset size in Section~\ref{the-effect-of-quantization}. Finally,
tfilte model is compiled with edgetpu compiler which is possible only
when all layers are supported by edge TPU. However, Coral also has a
limitation on input tensor size due to memory restrictions which
prevents compilation of the SCHP-parser with initial tensor size of 473.
Considering that height and width dimensions of input tensor must be
multiples of certain values, we change input tensor size to 286. For
comparison consistency we apply the described changes, i.e. layer
substitution and input tensor size modification, to all models. Then we
train parsers with ResNet-101, ResNet-18 \cite{Rp}, MobileNetV2
\cite{Mv2}, and OSNet \cite{OSN} backbones on LIP dataset and then use
inference masks obtained for re-ID datasets to calculate re-ID accuracy.
Comparing pure human parsing accuracy metrics such as mean intersection
over union (mIOU) is out of scope of this paper.

\begin{table}
  \caption{Compilation details of studied models depending on backbone.}
  \label{tab:2}
  \begin{center}
  \begin{tabular}{|c|c|c|c|c|}
    \hline
    \textbf{Parameter/backbone} & \textbf{ResNet-101} & \textbf{ResNet-18} & \textbf{MobileNetV2}
    & \textbf{OSNet} \\
    \hline
    Number of parameters, mln & 42.8 & 11.4 & 4.2 & 2.2 \\
    \hline
    Number of operations & 217 & 87 & 121 & 390 \\
    \hline
    Compilation time, s & 88.0 & 1.6 & 0.9 & 4.0 \\
    \hline
    Quantized model size, MiB & 65.13 & 12.27 & 2.73 &
    3.93 \\
    \hline
    Output size, MiB & 65.65 & 12.42 & 3.39 & 4.88 \\
    \hline
    Used memory*, MiB & 6.35 & 6.68 & 3.13 & 3.03 \\
    \hline
    Remaining memory**, KiB & 6.25 & 2.0 & 4536.32 & 1.5 \\
    \hline
    Off-chip memory***, MiB & 57.87 & 5.48 & 0 & 0.88 \\
    \hline
  \end{tabular}
  \end{center}
  * On-chip memory used for caching model parameters. \\
  ** On-chip memory remaining for caching model parameters. \\
  *** Off-chip memory used for streaming uncached model parameters.
\end{table}

\section{Experiments}
\label{experiments}

\subsection{Changes in re-ID accuracy depending on parser backbone
choice}
\label{changes-in-re-id-acc}

Table~\ref{tab:1} shows accuracy depending on parser backbone and model type:
original (PyTorch), quantized (tflite), and Coral (which is compiled
with edgetpu-complier). All backbones were first trained on GPU in
supervised learning fashion using LIP dataset for 100 epochs with
starting learning rate of 0.0035. Re-ID accuracy calculation requires
parsing masks for query and test images. For \emph{original} models,
parsing inference masks were obtained on GPU. For \emph{quantized}
models, parsers were quantized as discussed in Section~\ref{parser-mod} with 5000
images in quantization dataset and inference masks were obtained on GPU
using quantized parsers. For \emph{Coral} models, quantized parsers were
transferred to Coral and parsing masks were obtained on Coral. Re-ID
metrics were calculated on the same CPU using the analytical method.
Parsers were trained on a server with Intel i9-9900K 3.60GHz CPU with
64GB RAM and NVIDIA Quadro P5000 with 16GB RAM.

Conventional re-ID accuracy metrics, namely rank-1, rank-10, and mean
average precision (mAP), were used to evaluate the performance of
studied models \cite{reidoutlook}. All experiments were conducted on
Market1501 dataset consisting of 19732 test images and 3368 queries
\cite{markp}. It should be stressed that all parsers were trained on
original LIP dataset and not on re-ID datasets which makes them re-ID
dataset agnostic.

\subsection{Backbone performance comparison with Coral}
\label{backbone-performance-comparison-with-coral}

Models modified in accordance with Section~\ref{parser-mod} with input tensor size of
286 were successfully compiled with edgetpu complier. Table~\ref{tab:2} shows
that model size (in MiB) increases from MobileNetV2 to ResNet-101. An
interesting observation is that OSNet, while its size is slightly bigger
than that of MobileNetV2, has the smallest number of parameters and the
highest number of operations (even exceeding ResNet-101). This is
because OSNet has very complex structure with large number of
calculations despite of its small size and number of parameters, so it
has high compilation and inference time inferior only to those of
ResNet-101, as shown in Table~\ref{tab:3}. Nevertheless, the inference time
for OSNet is still sufficiently small for real-time applications. The
difference in time between the first and consequent inferences is a
common phenomenon related to the necessity to upload data and initialize
the model for the first inference. In should be noted that for tflite
models, both quantized and Coral, the reduction in inference time
reaches up to 20\%, whereas for PyTorch models on GPU it is even more
significant.

\begin{table}
  \caption{Inference time in ms.}
  \label{tab:3}
  \begin{center}
  \begin{tabular}{|c|c|c|c|c|}
    \hline
    \textbf{Type/backbone} & \textbf{ResNet-101} & \textbf{ResNet-18} & \textbf{MobileNetV2}
    & \textbf{OSNet} \\
    \hline
    Original (GPU) & 39 → 13 & 200 → 5 & 37 → 5 & 110 → 14 \\
    \hline
    Quantized (GPU) & 920 → 740 & 120 → 104 & 72 → 51 & 202 → 183 \\
    \hline
    Coral (TPU) & 300 → 280 & 50 → 40 & 32 → 26 & 58 → 49 \\
    \hline
  \end{tabular}
  \end{center}
  \small
  Arrows indicate change in time between first and consequent inferences
\end{table}

\subsection{The effect of quantization dataset size on parsing quality}
\label{the-effect-of-quantization}

\begin{figure} 
  \centering
  \includegraphics[scale=0.7]{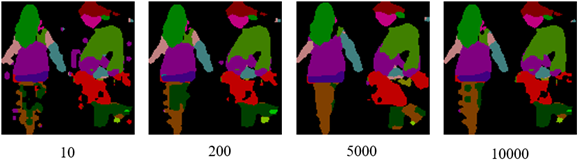}
  \caption{OSNet parsing masks for different numbers of images in
  quantization dataset.}
  \label{fig:1}
\end{figure}

\begin{figure} 
  \centering
  \includegraphics[scale=0.35]{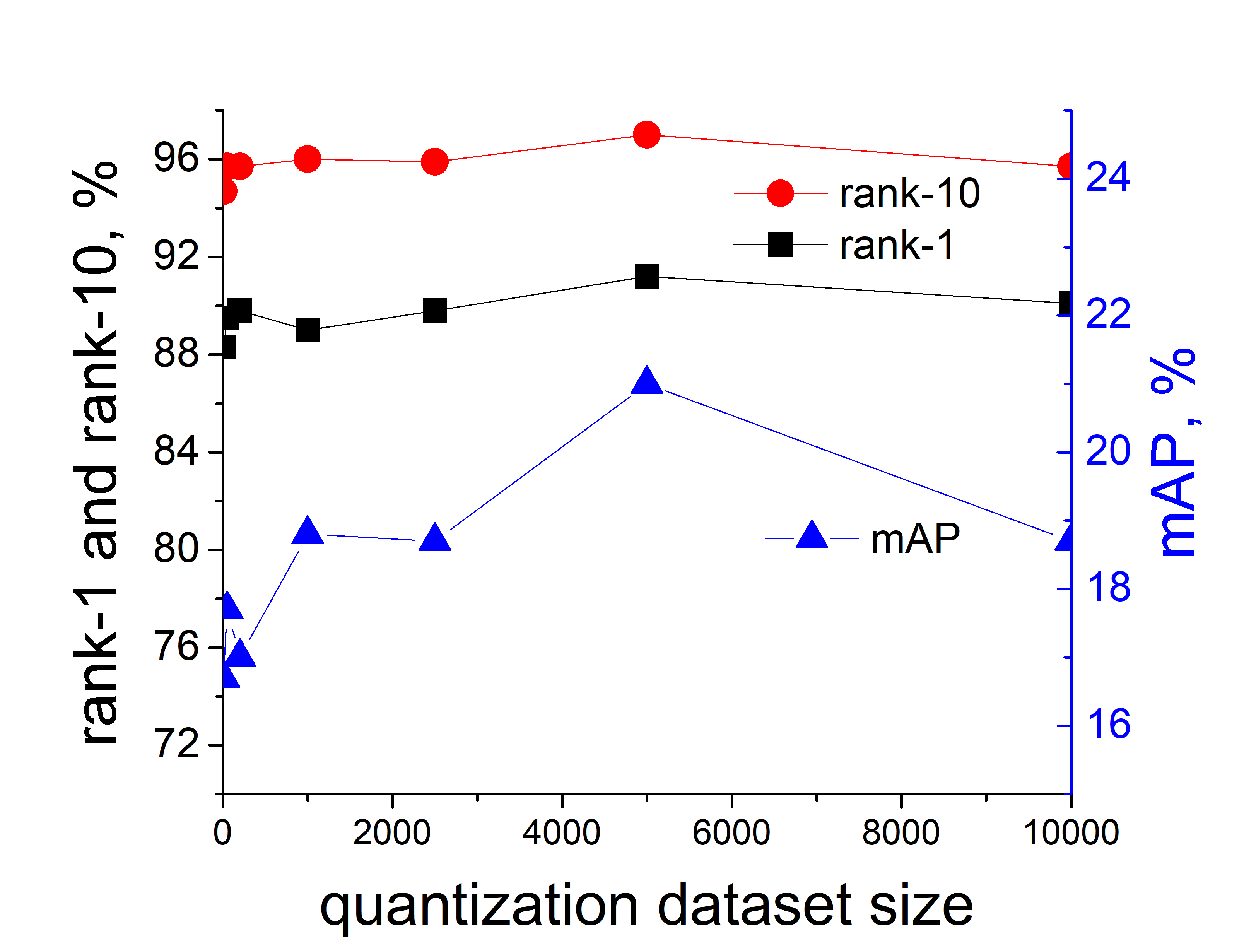}
  \caption{Changes in re-ID metrics quantized OSNet-based parser
  depending on quantization dataset size.}
  \label{fig:2}
\end{figure}

To study the effects of quantization dataset size on re-ID accuracy, we
performed experiments using a parser with OSNet backbone while changing
the number of images in the quantization dataset. We conduct experiments
for the dataset sizes of 10, 50, 200, 1000, 2500, 5000, and 10,000
images. Figure~\ref{fig:1} shows how parsing masks change for two sample
queries. In general, the increase in the dataset size leads to increase
in parsing quality: the number of misclassified regions decreases, noisy
predictions such as random class spots are vanishing, etc. Figure~\ref{fig:2}
shows how re-ID accuracy metrics change with dataset size. It shows
rapid increase in accuracy up to 200 images which then slowly increasing
up to 5000 images and then starting to decrease. It should be mentioned
that the correlation between changes in parsing masks and re-ID accuracy
is not straightforward, and significant changes in masks can correspond
to small changes in re-ID accuracy, as discussed in detail in the next
Section.

\section{Discussions}
\label{discussions}

\subsection{The effects of backbone change on human parsing quality and re-ID accuracy}
\label{the-effects-of-backbone-change}

Table~\ref{tab:1} shows that the highest accuracy for all three metrics is
achieved by the original version of ResNet-101. However, the overall
accuracy reduction due to backbone change is not significant -- within
3\% for rank-1, 2.5\% for rank-10, and 4.5\% for mAP. The decrease in
mAP is larger compared to other metrics, though lower mAP is a known
feature of studied method. At the same time the number of parameters,
size of the model, and inference speed decrease significantly, as shown
in Tables~\ref{tab:2} and ~\ref{tab:3}. This result shows that smaller parsers can indeed be
used with the proposed re-ID method with small accuracy reduction.

It should be noted that ResNet-101 parser used in this study has 1\%
higher rank-1/rank-10 accuracies and 1.2\% lower mAP accuracy than the
ResNet-101 SCHP parser used in \cite{combreid}. A possible reason for
this is that when parsing masks are extremely precise, the requirements
for similarity estimation also become stricter. At the same time the
effects of illumination and posture change also become more noticeable
for precise masks, and these effects are the most profound sources of
reidentification errors. Hence, having less precise parsing masks
loosens similarity criteria allowing the same person found in different
camera views be identified correctly. However, this also makes it more
likely that different people will be incorrectly identified as looking
similar which is reflected in reduced mAP.

Table~\ref{tab:1} also shows that for three backbones the accuracy decreases
after quantization. However, quantized and Coral versions of OSNet
outperform other backbones and almost reach the accuracy of the original
ResNet-101. This observation implies that this backbone is the most
promising for edge computing applications. The accuracy also only
slightly changes between quantized and Coral models despite of change of
the processor's architecture from GPU to TPU.

\begin{figure} 
  \centering
  \includegraphics[scale=0.53]{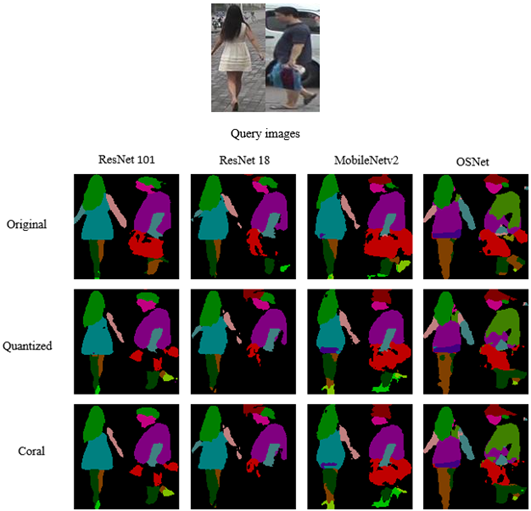}
  \caption{Comparison of parsing masks generated by different backbones
  for two sample queries.}
  \label{fig:3}
\end{figure}

Figure~\ref{fig:3} shows generated parsing masks for two sample queries using
studied backbones. The left query represents a simple case where a
person is clearly visible, and it is parsed relatively well by all
backbones. There are several cases where left hand is not parsed
correctly, and it sometimes disappears after quantization. However,
classification errors are more significant for OSNet since whereas the
parsing mask is overall correct, there are patches of mixed classes on
body and arms. It should be noted that in general the masks obtained
using the quantized parsers do not differ dramatically from those
obtained using the original networks. This is surprising because there
was a lot of transformations in network architecture associated with
the conversion which include quantization, i.e., a transition from
almost continuous floating point accuracy weights to their one-byte
discrete version.

The right query is more challenging since it includes partial occlusion
of person's legs by the bag which does not have a corresponding class in
LIP. Some parsers misclassified it as ``pants``, others excluded it from
the masks though it often resulted in exclusion of legs, too. These
masks are very noisy and include both region boundary errors and region
misclassification errors. There is a tendency for ResNets to lose some
regions after quantization, whereas changes are less dramatic for
MobileNetV2 and OSNet.

It is interesting that whereas parsing masks can change significantly,
the results in Table~\ref{tab:1} show that this fact has only small influence on
re-ID accuracy. This means that the proposed re-ID method is robust with
respect to parsing errors, since large changes in masks lead to little
changes in accuracy. In our opinion, there are two reasons for this.
First, parsers partially rely on colors when classifying image regions,
so misclassification adds similar colors to class histograms in our
method. Moreover, such regions often include shadows which are present
in every class, resulting in slight shifts in feature channel histograms
without causing significant disturbance. Second, the size of
misclassified areas is often small, so they have proportionally little
effect on feature channel histograms. Furthermore, semantically similar
classes are merged in our analysis which increases tolerance to
misclassification \cite{combreid}. However, errors that affect large
regions would negatively affect the accuracy. Nevertheless, the
discussed influence of parsing errors on re-ID accuracy clearly
distinguishes this study from pure DL-based human parsing experiments
where each pixel or area misclassification contributes to accuracy
reduction \cite{maskp}. Therefore, the use of smaller parser in pure
DL-based human parsing tasks should be undertaken with extra care,
whereas for the proposed re-ID method such operation is justified, as
shown by Table~\ref{tab:1} and Figure~\ref{fig:3}.

It should be noted that we trained parsers only for 100 epochs out of
150 epochs commonly used in human parsing studies \cite{ce2pp, schpp}. This
implies that the parsing results can be improved further, especially if
self-correction is applied \cite{schpp}. However, since improving human
parsing accuracy is not the main goal of this study, we present the
results for 100 epochs to further emphasize the point that the proposed
combined re-ID method is robust with respect to human parsing
inaccuracies.

\subsection{Application scenarios for edge computing Coral-based
re-ID}\label{application-scenarios}

Whereas re-ID metrics in Table~\ref{tab:1} were calculated on a PC, the
possibility of evaluating person similarity solely on Coral was also
verified. Coral performed the complete similarity evaluation pipeline by
parsing an image and using the obtained mask together with the image to
calculate feature vectors. We also performed a similarity ranking
experiment on Coral using 500 images from Market1501 which shows that
re-ID can in principle be realized on Coral. However, Coral has limited
RAM and CPU, so using it for large-scale re-ID with thousands or
millions of images is not possible. It should be noted that Coral still
performs sufficiently fast for real-time human parsing, so we propose
two possible application scenarios for Coral-based re-ID.

\subsubsection{ Coral as an edge computing image processing camera}
\label{coral-as-edge}

As we mention above, Coral is capable of real-time human parsing and
calculation of feature vectors. Thus, it can act as a useful edge
computing device that transmits preprocessed data to a server that
performs final operations for re-ID. In a conventional scheme such
server works with raw video streams from camera systems, which might
include hundreds and thousands of cameras. Therefore, parsing enormous
amounts of video streams and calculating features is an extremely
resource demanding task. On the contrary, only similarity score
calculation and ranking are to be performed when server obtains parsing
masks and feature vectors from the camera. It is easy to see that the
advantage of such scheme is proportional to the number of cameras in the
system, making it most suitable for video surveillance in smart
cities.

\subsubsection{Coral as a re-ID
detector}\label{coral-as-a-re-id-detector}

In addition to the above point, there is another role that Coral can
play. We have mentioned that the main limitation for the Coral-based
re-ID is memory, but Coral can store a small database to compare
observed people with. This is useful when a query person is, for
instance, expected to appear in certain area and Coral can send a
message to an operator when a match is found. In this scheme Coral
performs real-time comparison but does not necessarily store data other
than IDs of matches which can be transferred to the operator along with
masks and feature vectors. This is extremely useful for rapid response
in case of searching for lost people or looking for perpetrators. All
tested backbones provide the possibility to conduct this analysis in
real time.

\section{Conclusions}
\label{conclusions}

This paper studies the possibility of developing a compact and
computationally undemanding re-ID method and realizing it using Google
Coral Dev Board. We use a recently proposed combined re-ID method where
only human parsing section relies on DL and study the effects of
changing parser backbones. We show that re-ID accuracy decreases
slightly when substituting ResNet-101 backbone with ResNet-18,
MobileNetV2, and OSNet, whereas number of parameters and model size
exhibit more than ten- and twenty-fold reduction. We discuss necessary
modifications to DL models to make them Coral-compatible and apply said
modifications to all models. We quantize all models and transfer them to
Coral with re-ID accuracy decreasing slightly for ResNet-101, ResNet-18,
and MobileNetV2 compared to GPU. However, both quantized and Coral
versions of OSNet show higher accuracy than its GPU version also being
the highest among all transferred models. This makes OSNet the most
promising candidate for edge computing re-ID applications. We also show
that the reduction in parsing mask accuracy is greater than the
reduction in re-ID accuracy when backbones change and argue that this
indicates robustness of our method with respect to parsing errors. We
also discuss the effects of quantization dataset size on re-ID accuracy
and suggest range of suitable sizes. Finally, we show that all models
can perform in real-time and discuss how Coral can act as edge computing
parser or perform complete re-ID cycle.

\section*{Acknowledgement}
\label{acknowledgement}

Authors would like to thank their Kryptonite colleagues Dr Igor Netay
for fruitful discussions and Vasilii Dolmatov for his assistance in
problem formulation, choice of methodology, and supervision.


\bibliographystyle{IEEEtran}
\bibliography{IEEEabrv,ref}